\documentclass[letterpaper, 10 pt, conference]{ieeeconf}  

\IEEEoverridecommandlockouts                              

\usepackage{times}

\usepackage{algorithm}
\usepackage[noend]{algcompatible}

\usepackage{amsmath}
\usepackage{amsfonts}
\usepackage{breqn}
\usepackage{multirow,hhline,graphicx,array}
\graphicspath{ {./images/} }
\usepackage{multicol}
\usepackage[bookmarks=true]{hyperref}
\usepackage{subfiles}
\usepackage{caption}
\usepackage{float}
\usepackage[flushleft]{threeparttable}

\usepackage{xcolor}

\title{\LARGE \bf
ROOAD: RELLIS Off-road Odometry Analysis Dataset
}

\author{George Chustz$^{1}$ and Srikanth Saripalli$^{1}$
\thanks{$^{1}$J. Mike Walker '66 Department of Mechanical Engineering, 
Texas A\&M University, College Station, TX 77843, USA
        {\tt\small \{gchustz, ssaripalli\}@tamu.edu}}%
}

\begin{document}

\maketitle
\thispagestyle{empty}
\pagestyle{empty}

\begin{abstract}
The development and implementation of visual-inertial odometry (VIO) has focused on structured environments, but interest in localization in off-road environments is growing. In this paper, we present the RELLIS Off-road Odometry Analysis Dataset (ROOAD) which provides high-quality, time-synchronized off-road monocular visual-inertial data sequences to further the development of related research. We evaluated the dataset on two state-of-the-art VIO algorithms, (1) Open-VINS and (2) VINS-Fusion. Our findings indicate that both algorithms perform 2 to 30 times worse on the ROOAD dataset compared to their performance in structured environments. Furthermore, OpenVINS has better tracking stability and real-time performance than VINS-Fusion in the off-road environment, while VINS-Fusion outperformed OpenVINS in tracking accuracy in several data sequences. Since the camera-IMU calibration tool from Kalibr toolkit is used extensively in this work, we have included several calibration data sequences. Our hand measurements show Kalibr's tool achieved \(\pm\)1\textdegree~for orientation error and \(\pm\)1 mm at best (x- and y-axis) and \(\pm\)10 mm
(z-axis) at worse for position error in the camera frame between the camera and IMU. This novel dataset provides a new set of scenarios for researchers to design and test their localization algorithms on, as well as critical insights in the current performance of VIO in off-road environments.

ROOAD Dataset: github.com/unmannedlab/ROOAD
\end{abstract}

\section{Introduction}

Localization is one of the fundamental problems in robotics. The development and analysis of localization algorithms requires detailed and carefully acquired datasets from a variety of scenarios and locations. There are many high-quality datasets which are publicly available; however, the majority of these do not provide off-road data. The research and development of autonomous vehicles is primarily focused on urban environments\cite{Tan2020,Geiger2013IJRR,Caesar2019,Cordts2016,Yu2018,Neuhold2017,Geyer2020}, highways, warehouses, indoor locations\cite{sturm12iros,Delmerico19icra,Burri25012016}, simulations\cite{WangSCJDZLMTWLX19}, and other developed areas.

Off-road environments provide a different set of challenges than urban environments. While there are far less pedestrians, vehicles, and rules, there are insects, animals, rocks, and plants. Ground vehicle dynamics change when driving over clay, dirt, sand, or mud. Obstacles should be avoided when on roads, but off-road, some are navigable by the vehicle. Haze and cloud cover can change the lighting conditions and visibility dramatically. Thus, there are different LIDAR and camera features to track, and the variety in texture of driving surfaces provides different excitation profiles for inertial sensors.

\begin{figure}
  \centering
  \includegraphics[width=0.48\textwidth]{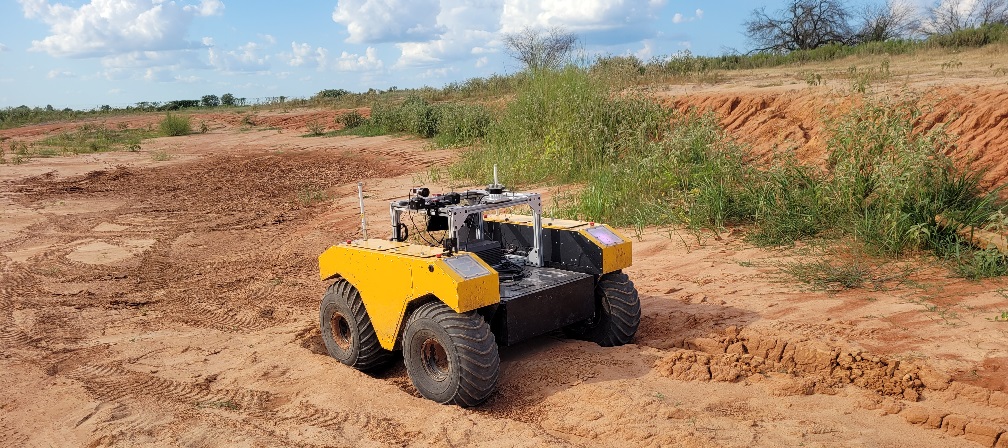}
  \caption{Image of Clear Path's Warthog Platform\cite{warthog} at the testing site. This location is a dried mud pit within the Up-Down path.}
  \label{fig:terrain}
\end{figure}

Visual-Inertial Odometry (VIO) is one such set of algorithms which estimate the local orientation and position of a vehicle using an Inertial-Measurement Unit (IMU) and cameras. VIO datasets need to be carefully acquired with special attention to the time stamping of images and IMU readings. While cameras and IMUs are relatively inexpensive sensors, acquiring accurate time stamps and performing time synchronization can be expensive and labor intensive.

In this work, we introduce RELLIS Offroad Odometry Analysis Dataset (ROOAD) to contribute time-synchronized monocular VIO data to further the development of this field. The data sequences are recorded at a desert off-road testing field surrounding a sinkhole at the Texas A\&M University System RELLIS Campus (Fig.~\ref{fig:map}), using ClearPath's Warthog\cite{warthog} platform (Fig.~\ref{fig:terrain}) and IEEE-1588 (PTP)\cite{PTP2016} to accurately time stamp the raw data from the camera. The performances of OpenVINS and VINS-Fusion are evaluated with respect to real-time kinematic positioning (RTK) GPS groundtruth, and the performance of the camera-IMU extrinsics calibration tool Kalibr is measured using the precision hatching on the camera slide (shown in Fig.~\ref{fig:vio_sensors}). The major contributions of this work are as follows:

\begin{itemize}
    \item We release six time stamped visual, inertial, and ground truth data sequences acquired while traversing off-road at the RELLIS testing grounds.
    \item We demonstrate the performance of the state-of-the-art VIO implementation OpenVINS and VINS-Fusion on our off-road traversal data sequences.    
    \item We validate the efficacy of the Kalibr camera-IMU cross calibration tool through the release and analysis of 11 additional calibration datasets with various relative horizontal position and yaw orientations between the camera and IMU.
\end{itemize}

\section{Related Work}
\subsection{Datasets}

There is a large collection of publicly available autonomous vehicle traversal datasets for researchers to develop their work. Some examples of high quality multi-modal urban datasets include KITTI\cite{Geiger2013IJRR} and NuScenes\cite{Caesar2019}, and other datasets focused only on images and point clouds for semantic segmentation (Cityscapes\cite{Cordts2016}, BDD100K\cite{Yu2018}, Mapillary Vistas\cite{Neuhold2017}, Toronto-3D\cite{Tan2020}, and A2D2\cite{Geyer2020} which cannot be used for localization). Many VIO-centric datasets are acquired indoors such as TUM\cite{sturm12iros}, UZH-FPV\cite{Delmerico19icra}, and EuRoC\cite{Burri25012016}. Lastly, there are simulation-based data sets such as Tartan AIR\cite{tartanair2020iros}.

\begin{table}[]
\centering
\vspace{10pt}
\captionsetup{margin={0pt,0pt}}
\caption{Publicly available off-road datasets comparison.}
\resizebox{0.48\textwidth}{!}{%
\begin{tabular}{c|c|c|c|c|c|}
\cline{2-6}
                                  & ROOAD      & RELLIS    & Frieburg & YCOR     & RUGD    \\
                                  &            & -3D       & Forest   &          &         \\ \hline
\multicolumn{1}{|c|}{Best}        & 30 Hz      & 10 Hz     & 20 Hz    & 10 Hz    & 15 Hz   \\
\multicolumn{1}{|c|}{Camera}      & Grayscale  & RGB       & RGB      & RGB      & RGB     \\
\multicolumn{1}{|c|}{}            & 1920x1200  & 1920x1200 & 1024x768 & 1024x544 & 688x550 \\ \hline
\multicolumn{1}{|c|}{IMU}         & 400Hz      & 100 Hz    & X        & X        & X       \\ \hline
\multicolumn{1}{|c|}{Time Sync}   & Camera     & Cameras   & X        & X        & X       \\
\multicolumn{1}{|c|}{}            & + IMU      & + Lidar   &          &          &         \\ \hline
\multicolumn{1}{|c|}{Position}    & RTK GPS    & X         & X        & X        & X       \\
\multicolumn{1}{|c|}{Groundtruth} & +  heading &           &          &          &         \\ \hline
\end{tabular}%
}
\label{tbl:compare}
\end{table}


Notably there is less availability in off-road datasets in literature. Publicly available high-quality off-road datasets include RUGD\cite{RUGD2019IROS}, YCOR\cite{Maturana2018}, RELLIS-3D\cite{jiang2020icra}, and Deepscene's Freiburg Forest dataset\cite{valada16iser}. Freiburg Forest, YCOR, and RUGD only provide camera data which limits the use of these datasets in VIO applications due to lack of inertial data. RELLIS-3D provides a multi-modal dataset consisting of a RGB camera, stereo camera, LIDAR, DGPS, and IMU. While this dataset from the diverse sensor suite of RELLIS-3D is a valuable contribution to the community, the lack of IMU synchronization, slower sensor sample rates, and absence of position groundtruth make testing and evaluating VIO algorithms difficult. We have collected the off-road data sequences exhibited in ROOAD explicitly for VIO applications thereby addressing the limitations of the aforementioned datasets.

\subsection{Algorithms}

There are several publicly available, state-of-the-art Visual(-inertial) Odometry implementations such as OKVIS\cite{leutenegger2015keyframe}, R-VIO\cite{huai2019robocentric}, ROVIO\cite{Bloesch2017}, S-MSCKF\cite{sun2018robust}, and ICE-BA\cite{liu2018ice}. While all of these implementations have advanced this research area, we focus on localization in off-road environments and evaluate the recent VIO implementations, VINS-Fusion\cite{qin2019general} and OpenVINS\cite{Geneva2020ICRA} in these scenarios. VINS-Fusion utilizes optimization-based estimation whereas OpenVINS uses a Kalman Filter-based approach. These implementations have been extensively tested in urban and indoors environments, and in this paper their performance is evaluated in off-road scenarios.

\section{Sensor Setup and Calibration}
\subsection{Sensors}

\begin{figure}
  \centering
  \includegraphics[width=0.5\textwidth]{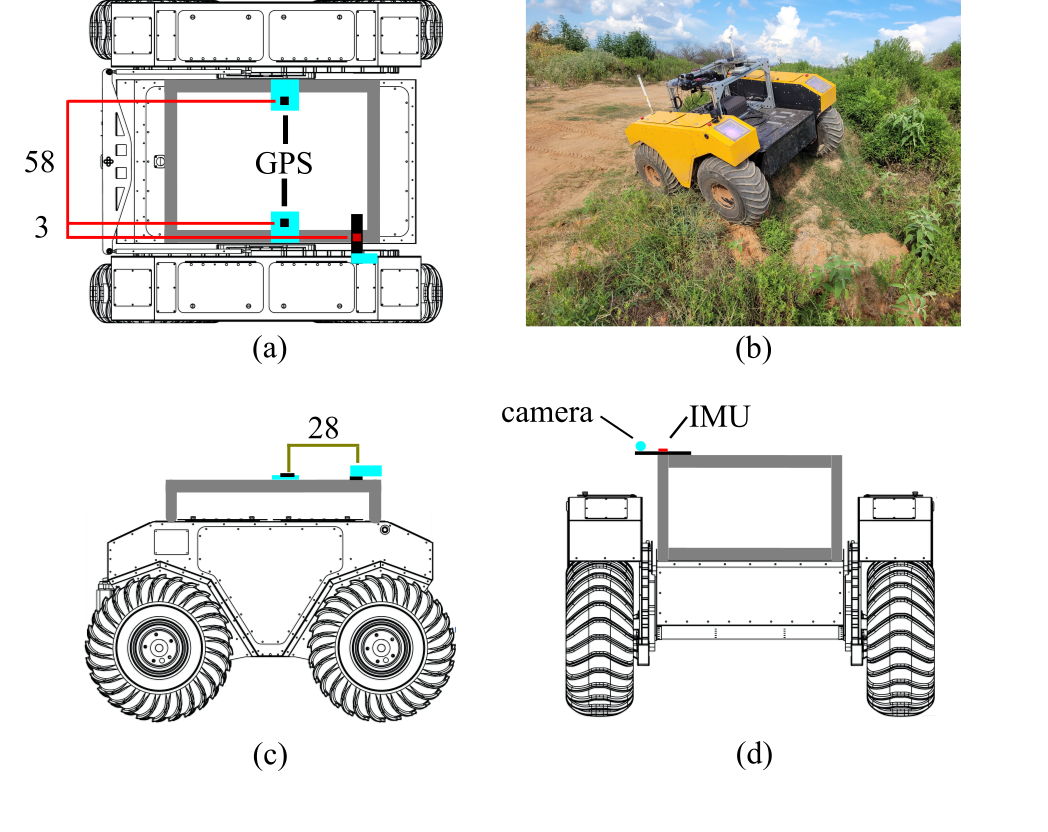}
  \caption{Clearpath's Warthog Platform\cite{warthog} Configuration. Illustration of the dimensions and mounting positions of the sensors with respect to the Warthog's top (a), side (c), and front (d) view. An image of the Warthog traversing a hill is included (b).  (Units: cm)}
  \label{fig:vio_sensors}
\end{figure}
\vspace{-5pt}
The acquisition of VIO-centric data necessitated an IMU with adequate noise density characteristics and a camera which allows for time stamping of images. The sensor set is labeled with dimensions in Fig.~\ref{fig:vio_sensors} and consists of the following sensors:
\begin{itemize}
  \item 1 \(\times\) B/W Camera: Basler acA1920-50gc camera with 8 mm/F1.8 Edmund Optics lens, resolution 1920x1200, 30 Hz, 1 ms exposure
  \item 1 \(\times\) IMU: Vectornav VN-300 INS at 400 Hz (GPS disabled)
  \item 1 \(\times\) GPS: Ardusimple Differential Heading with RTK kit, 10 Hz, local Ardusimple Basestation, all ZED-F9P GPS Modules
\end{itemize}

For the data collection, a computer with Power Over Ethernet (POE) and PTP capable network interface card is mounted to the top of the Warthog. The calibration, OpenVINS and VINS-Fusion implementations, and Kalibr evaluations are performed offline on a separate computer. Both computers run Ubuntu 20.04 (64 bit) and ROS Noetic for data collection and evaluation.

\subsection{Synchronization}

Using the Linux PTP package for Ubuntu, a boundary clock is established using the system's clock as the grandmaster and the network interface as the slave. The Basler Pylon Camera ROS driver was modified\footnote{https://github.com/unmannedlab/pylon-ros-camera} to use PTP time stamps from the camera itself. Unfortunately, the Vectornav IMU does not support PTP time synchronization, and the IMU's and system's clocks diverge over time. A one-dimensional Kalman filter\cite{trigger2015} is used\footnote{https://github.com/unmannedlab/vectornav/tree/feature/resync\_imu} to correct the discrepancy between the IMU and system clock. The time synchronization is verified prior and during every data collection run by monitoring the end-to-end latencies of the IMU and camera time stamps with respect to the system time.

\subsection{Camera Calibration}
Both OpenVINS, VINS-Fusion, and Kalibr support the pin-hole projection and radial-tangential distortion models. The pin-hole projection model assumes that the camera is projecting three-dimensional points onto an image plane through a singular point at the camera's aperture, and the radial tangential distortion model assumes that the lens distortion and sensor-lens misalignment can be modeled using polynomial approximation. The pinhole camera model matrix and the radial-tangential distortion parameters are estimated using the Kalibr toolbox. For the benchmarks and evaluations, we assume that the pin-hole model for the camera calibration and radial-tangential distortion model are adequate.  

\subsection{IMU Calibration}
For the IMU calibration, the Kalibr Allan toolbox determines the noise density and uses Allan Variance to estimate the random walk of the angle rate gyroscope and linear accelerometer. The data used to calculate the Allan Variance is approximately six-hours of stationary IMU data at 400 Hz. The linear acceleration noise density \(\sigma_{a}\) and random walk \(\sigma_{aw}\) is estimated to be \(1.1\times10^{-4} m/s^2/\sqrt{Hz}\) and \(3.2\times10^{-5} m/s^3/\sqrt{Hz}\), respectively, and the rate gyroscope noise density \(\sigma_{g}\) and random walk \(\sigma_{gw}\) is \(6.5\times10^{-5} rad/s/\sqrt{Hz}\) and \(5.0\times10^{-6} rad/s^2/\sqrt{Hz}\), respectively.
\subsection{IMU \& Camera Calibration}

The IMU-Camera cross calibration estimates the translation and orientation transform between the IMU and camera as well as the potential time offset between the camera and IMU. This cross calibration is performed using Kalibr calibrate camera-IMU tool. The estimate is verified using hand measured values and the efficacy of Kalibr's estimation ability is tabulated in Tables \ref{tbl:kalibr_pos} and \ref{tbl:kalibr_angle}.

\subsection{Ground Truth GPS and Heading}

Ardusimple's simpleRTK2B and simpleRTK2Blite breakout boards\cite{ardusimple} are used to acquire ground truth differential RTK GPS and heading. For this dataset, RTK GPS coordinates consisting of latitude, longitude, height are required to analyze the estimated position, and heading from RTK differential GPS is recorded to analyze orientation estimation performance. The base station is a singular GPS receiver, which surveys its location using the assumption that its position is stationary. The base station is able to transmit corrections to the differential GPS pair over 900MHz telemetry radios included in the Ardusimple kit. A single simpleRTK2B board acts as an RTK base station while a simpleRTK2B and a simpleRTK2Blite are mounted on the Warthog with GPS antennas placed as shown in Fig.~\ref{fig:vio_sensors}. The Warthog's simpleRTK2Blite board is in a "moving base station" configuration which receives RTK corrections from the stationary base station. The Warthog's simpleRTK2B board is in a "rover" configuration which determines the craft's GPS position and heading based on the relative heading of the simpleRTK2Blite's GPS antenna.
%

\section{Dataset}
\begin{figure}
  \centering
  \includegraphics[width=0.5\textwidth]{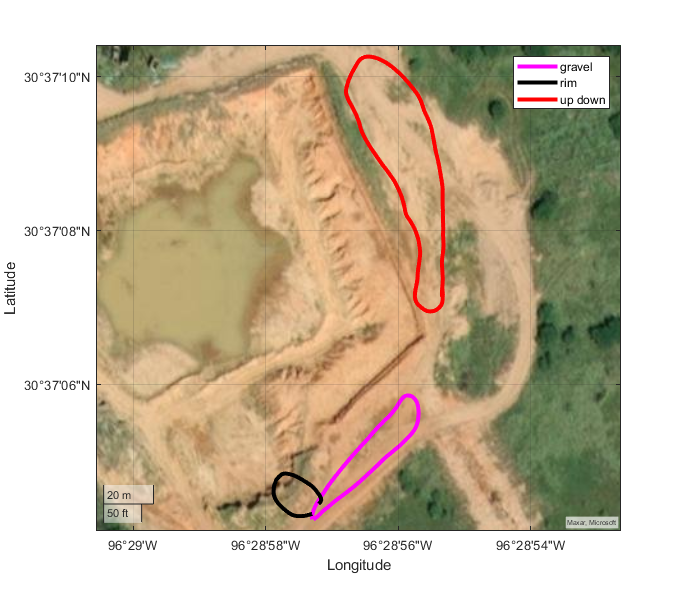}
  \caption{Satelite imagery of the testing ground with the groundtruth path coordinates overlaid.}
  \label{fig:map}
\end{figure}

ROOAD includes six sequences of Visual-Inertial data along three paths (shown in Fig.~\ref{fig:map}). The location presents desert conditions including shrubs, hills, and various elevations of terrain. The terrain of the testing field surrounds a sinkhole at the RELLIS Campus. The first path, "gravel," is acquired by driving on a stretch of gravel, then driving off-road adjacent to the path back to the starting point. The second path, "rim," is focused on a path which approaches the edge of the sinkhole before turning and returning back to the starting location. The final path, "updown," includes traversal along a dirt path, descent onto an eroded plateau, and return to the starting path. Each of these paths are recorded once at around 10 AM and again at around 7 PM local time to capture different lighting scenarios. The image, IMU, and GPS message counts for the individual datasets are given in Table~\ref{tbl:datasetinfo} and the maximum achieved excitations are shown in Table~\ref{tbl:datasetmax}. ROOAD contains some large excitations in excess of 3g's and large angle rates measured at the IMU to test the capabilities of VIO implementations. ROOAD is comprised of over 6,000 GPS waypoints, 20,000 images, and 250,000 IMU readings across its traversal data sequences to provide visual-inertial algorithm developers a new test-bed for their work.

\vspace{-5pt}      
\begin{table}[h]
\centering
\vspace{10pt}
\captionsetup{margin={0pt,0pt}}
\caption{ROOAD data sequences overview.}
\begin{tabular}{c|c|c|c|c|c}
\hline
Data     & Dist.     & Images   & IMU          & RTK GPS  & Time  \\
Sequence & (m)          &          & Messages     & Messages & Period \\
\hline
rt4\_gravel  & 135.8  & 3433        & 45765       & 1144     & evening \\
\hline
rt4\_rim     & 51.9   & 2085        & 27804       & 695      & evening \\
\hline
rt4\_updown  & 241.9  & 5364        & 71519       & 1787     & evening \\
\hline
rt5\_gravel & 132.4   & 2980        & 39740       & 994      & morning  \\
\hline
rt5\_rim    & 51.6    & 2257       & 30098        & 748       & morning \\
\hline
rt5\_updown & 221.7   & 4447       & 59290        & 1481    & morning  \\
\hline
\end{tabular}

\label{tbl:datasetinfo}
\vspace{-10pt}

\end{table}

\begin{table*}[h]
\centering
\vspace{10pt}
\captionsetup{margin={0pt,0pt}}
\caption{Maximum linear and angular excitation per dataset. The maximum excitations are shown in bold. (Units: m/s\textsuperscript{2} and rad/s)}
\begin{tabular}{c|c|c|c|c|c|c}
\hline
Data Sequence          & \(\ddot{x}_{max}\) & \(\ddot{y}_{max}\) & \(\ddot{z}_{max}\) & \(\omega_{x,max}\) & \(\omega_{y,max}\) & \(\omega_{z,max}\)       \\
\hline
rt4\_gravel            & -22.6 & 7.0 & 7.5 & 0.4 & -0.7 & -0.6 \\
\hline
rt4\_rim               & -10.5 & 6.4 & -4.9 & 0.5 & -0.6 & 0.5 \\
\hline
rt4\_updown            & \textbf{31.9} & -11.2 & -15.5  & 0.8  & 1.4 & -0.6 \\
\hline
rt5\_gravel            & -27.0  & -13.7  & 11.9  & 0.9  & 1.0 & -0.6 \\
\hline
rt5\_rim               & -11.0 & 7.5 & -5.1  & 0.8 & 0.6  & 0.4 \\
\hline
rt5\_updown            & -30.8   & \textbf{-20.5} & \textbf{-21.3} & \textbf{2.1} & \textbf{1.4} & \textbf{ -0.8} \\
\hline
\end{tabular}

%
\label{tbl:datasetmax}
\vspace{-10pt}

\end{table*}

\vspace{0.01\textheight}

Additionally, there are 11 camera-IMU cross calibration data sequences provided. For each of the sequences in this group, the camera is either translated along the IMU's x-axis while the rotation is held constant or rotated about the IMU's z-axis while the translation is held constant. Since Kalibr's calibration tools are used widely within the research community, the validation of these tools would provide useful feedback and consideration for those who are using Kalibr or developing counterparts. The Kalibr calibration analysis shows adequate repeatability and accuracy among the different configurations tested.   

\section{Benchmarks, Evaluation Metrics and Experiments}
\begin{table*}[h]
\centering
\vspace{10pt}
\captionsetup{margin={0pt,0pt}}
\caption{Median relative pose error of OpenVINS (OV) \& VINS-Fusion (VF) for the provided datasets. Bolded values are the better of the two algorithms of each metric of each dataset.}
\begin{tabular}{c|c|c|c|c|c|c}
\hline
File Name              & 8m OV                           & 8m VF                         & 24m OV               & 24m VF               & 40m OV                & 40m VF        \\
\hline
rt4\_gravel            & \textbf{1.9\textdegree}/0.8m   & 2.5\textdegree/\textbf{0.7m} & \textbf{2.7\textdegree}/2.4m & 3.1\textdegree /\textbf{1.6m} & 5.3\textdegree/3.6m & \textbf{4.4\textdegree}/\textbf{2.7m} \\
\hline
rt4\_rim               & \textbf{6.0\textdegree}/\textbf{1.5m} & 6.8\textdegree/4.4m & \textbf{9.6\textdegree}/\textbf{3.0m} & 14.0\textdegree/19.4m & \textbf{9.4\textdegree}/\textbf{1.6m} & N/A / N/A\textsuperscript{1} \\
\hline
rt4\_updown            & \textbf{3.5\textdegree}/0.9m & 3.7\textdegree/\textbf{0.8m} & \textbf{5.7\textdegree}/2.5m & 6.0\textdegree/\textbf{1.9m} & \textbf{8.1\textdegree}/3.7m & 8.5\textdegree/\textbf{2.7m} \\
\hline
rt5\_gravel            & \textbf{2.2\textdegree}/\textbf{0.6m} & 2.7\textdegree/0.8m & \textbf{3.4\textdegree}/\textbf{1.8m} & 4.7\textdegree/2.0m & \textbf{6.0\textdegree}/\textbf{2.8m} & 7.8\textdegree/3.1m \\
\hline
rt5\_rim               & \textbf{5.8\textdegree}/\textbf{1.3m} & 6.5\textdegree/247.6m & \textbf{9.7\textdegree}/\textbf{2.7m} & 11.8\textdegree/736.5m & \textbf{8.2\textdegree}/\textbf{2.4m} & N/A / N/A\textsuperscript{1} \\
\hline
rt5\_updown            & \textbf{3.6\textdegree}/0.8m & 3.8\textdegree/\textbf{0.8m} & \textbf{6.4\textdegree}/1.8m & 6.7\textdegree/\textbf{1.8m} & \textbf{9.1\textdegree}/2.4m & 9.5\textdegree/\textbf{2.3m} \\
\hline
\end{tabular}

\begin{tablenotes}
    \centering
    \small
    \item \textsuperscript{1} Divergent values caused errors when calculating the relative pose error.
    
\end{tablenotes}
\label{tbl:datasetrpe}
\vspace{-10pt}

\end{table*}

\begin{table}[h]
\centering
\vspace{10pt}
\captionsetup{margin={0pt,0pt}}
\caption{Results of Kalibr Horizontal Trials. All other rotational and translational modes are fixed. The expected y and z relative measurements are 75 mm and -30 mm, respectively. (Units: mm)}
\begin{tabular}{c|c|c|c}
\hline
Slider Measurement & X & Y & Z  \\
\hline
120         & 119       & 68   & -66 \\
\hline
90        & 89        & 67   & -68 \\
\hline
60        & 59        & 67   & -71 \\
\hline
-70       & -69       & 70   & -85  \\
\hline
-100       & -99       & 68   & -86  \\
\hline
-130       & -129      & 71   & -90  \\
\hline
\end{tabular}

\label{tbl:kalibr_pos}
\vspace{-10pt}

\end{table}

\begin{table}[h]
\centering
\vspace{10pt}
\captionsetup{margin={0pt,0pt}}
\caption{Results of Kalibr Yaw Rotation Trials. All other rotational and translational modes are fixed. The measured x and y Euler Angles are -90\textdegree~and 0\textdegree, respectively. (Units: degrees)}
\begin{tabular}{c|c|c|c}
\hline
Measured Z-Angle & X & Y & Z  \\
\hline
-125         & -90.3      & -0.1   & -125.3 \\
\hline
-110        & -90.2        & -0.4   & -109.5 \\
\hline
-95        & -90.4        & -2.9   & -92.6 \\
\hline
-80       & -89.9       & -0.9   & -79.5  \\
\hline
-65       & -89.7       & -1.0   & -64.9  \\
\hline
\end{tabular}

\label{tbl:kalibr_angle}
\vspace{-10pt}

\end{table}

\begin{figure}[!b]
  \centering
  \includegraphics[width=0.5\textwidth]{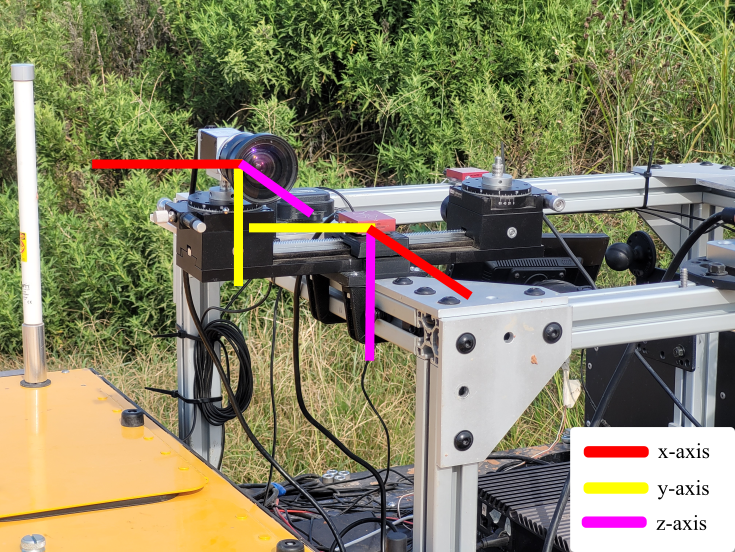}
  \caption{Warthog Sensor Orientation. Illustration of the co-ordinate axes of the IMU and camera. The camera's sensor is located at approximately 3, 13, and -7.5 cm on the x, y, and z-axes of the IMU co-ordinate system, respectively.}
  \label{fig:camimuori}
\end{figure}
\vspace{-5pt}

\begin{figure*}[h]
\centering
\vspace{10pt}
\captionsetup{margin={0pt,0pt}}

\begin{tabular}{ccc}

 (a) & (1) OpenVINS             & (2) VINS-Fusion \\

 (b) &  
 \includegraphics[width=0.45\textwidth]{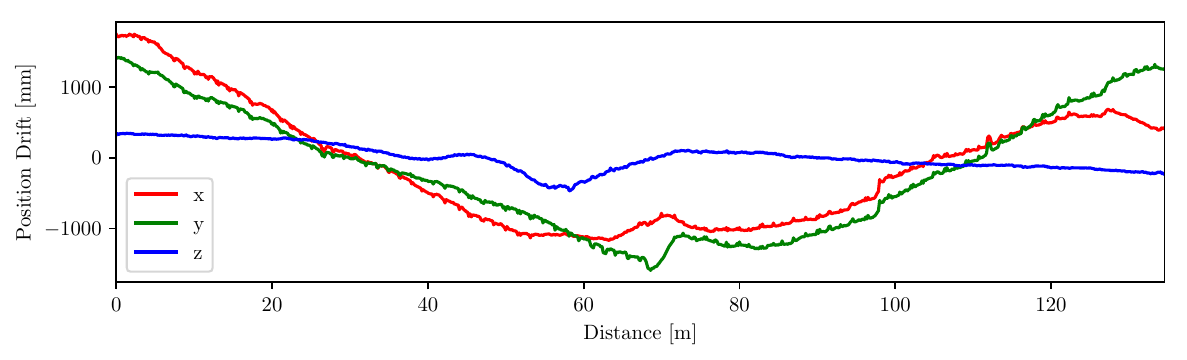} &
 \includegraphics[width=0.45\textwidth]{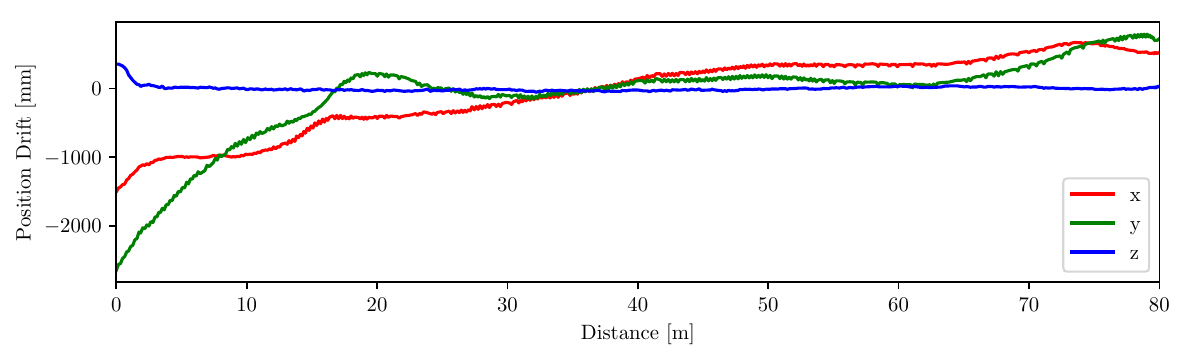} \\

 (c) & 
 \includegraphics[width=0.45\textwidth]{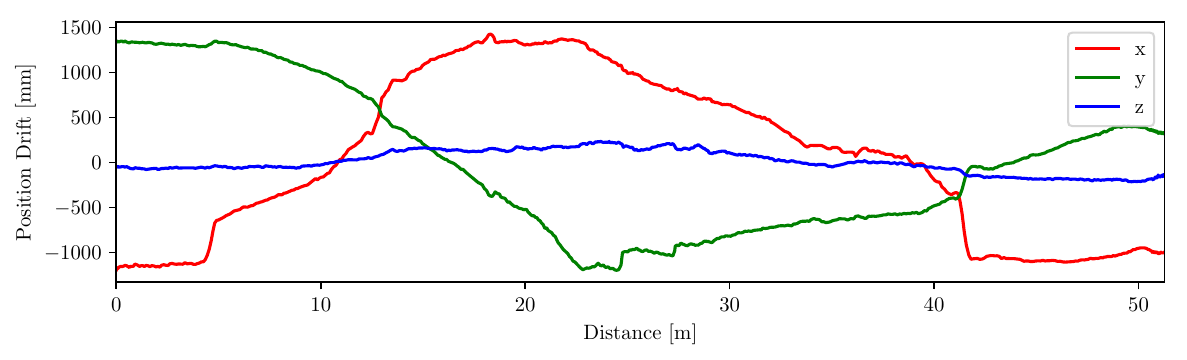} &
 \includegraphics[width=0.45\textwidth]{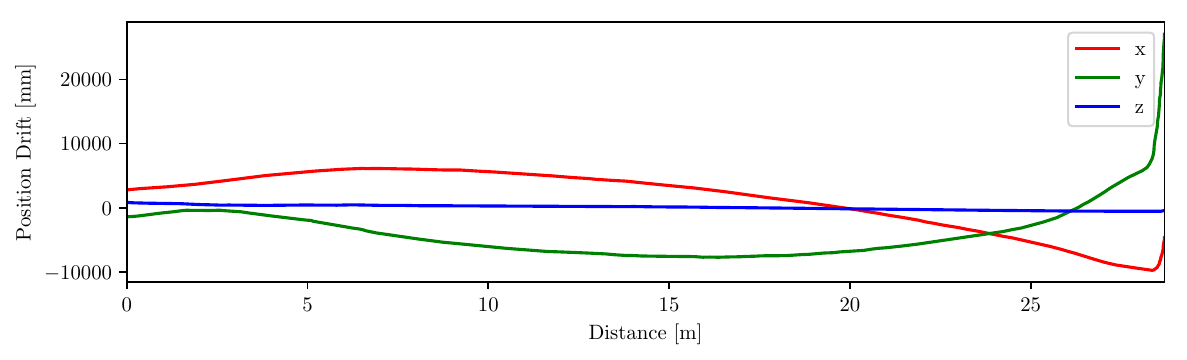} \\

 & 
 \includegraphics[width=0.45\textwidth]{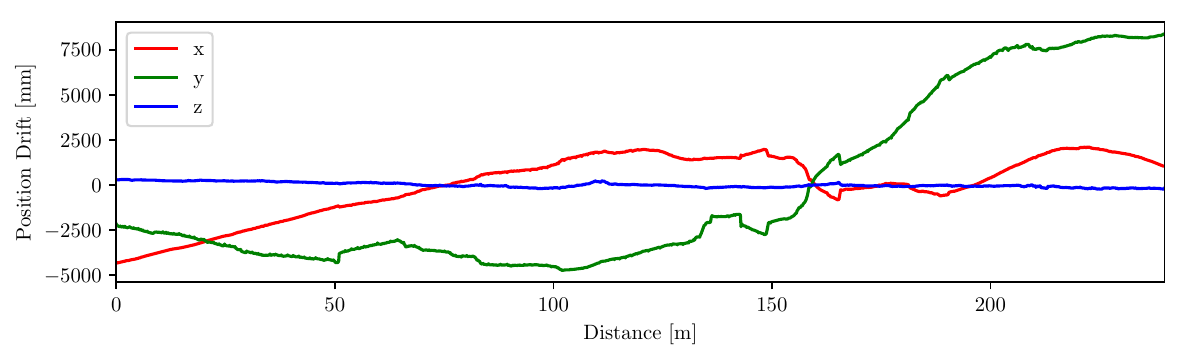} &
 \includegraphics[width=0.45\textwidth]{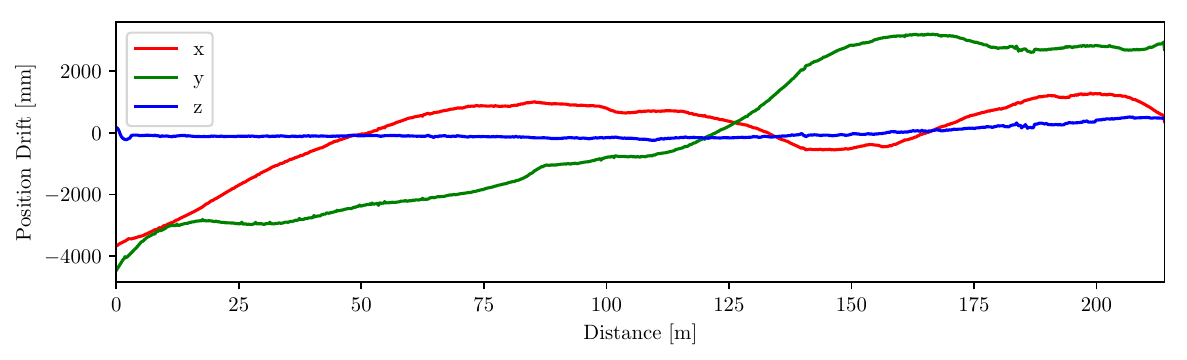} \\

\end{tabular}


\vspace{-10pt}
\caption{Comparison of translation errors. (a) represents the gravel path, (b) represents the rim path, and (c) represents the up-down path of the rt4 (evening) data sequences. (Units: mm/m)}
\label{fig:translationerror}
\end{figure*}

\begin{figure*}[h]
\centering
\vspace{10pt}
\captionsetup{margin={0pt,0pt}}

\begin{tabular}{ccc}

 (a) & (1) OpenVINS             & (2) VINS-Fusion \\

 (b) &  
 \includegraphics[width=0.35\textwidth]{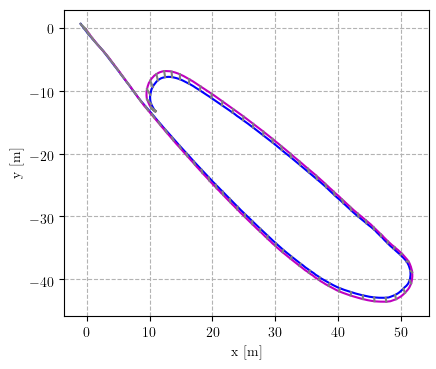} &
 \includegraphics[width=0.45\textwidth]{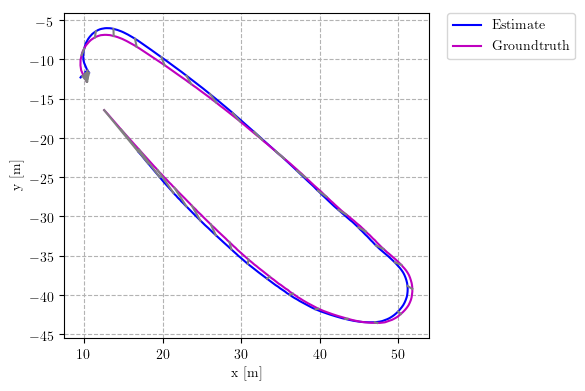} \\

(c) & 
 \includegraphics[width=0.35\textwidth]{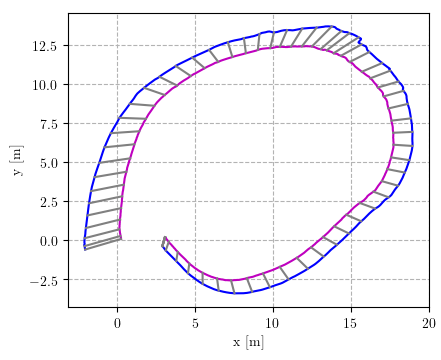} &
 \includegraphics[width=0.25\textwidth]{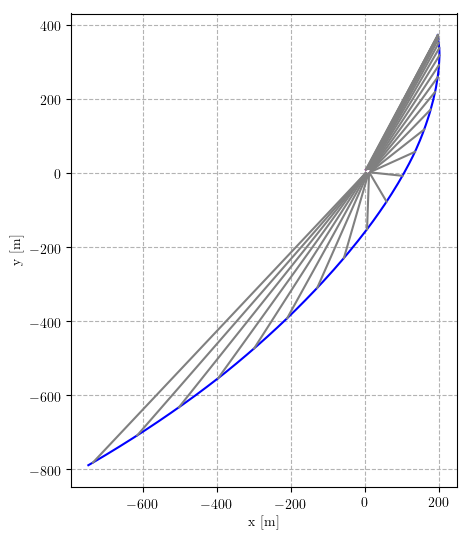} \\

 & 
 \includegraphics[width=0.35\textwidth]{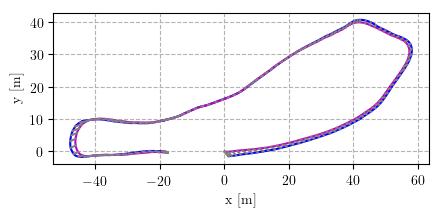} &
 \includegraphics[width=0.35\textwidth]{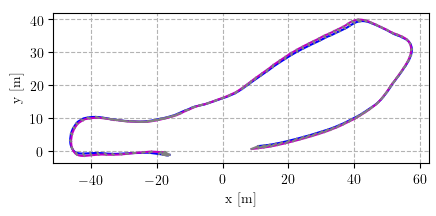} \\

\end{tabular}



\vspace{-10pt}
\caption{Comparison of trajectory estimates aligned to ground truth GPS of the (a) gravel path, (b) rim path, and (c) the up-down path for OpenVINS (1) and VINS-Fusion (2). The rim path was the most difficult for OpenVINS and VINS-Fusion was highly divergent on it. The gray lines between the estimated trajectory and the ground truth trajectory show the error after the trajectories have been aligned. (Units: m)}
\label{fig:trajectoryerrortopdownupdown}
\end{figure*}

\vspace{0.01\textheight}
The analysis of the OpenVINS and VINS-Fusion trajectories consists of comparing the groundtruth trajectories of the data sequences with the aligned VIO estimated trajectories. The groundtruth RTK differential GPS is capable of measuring latitude, longitude, height, and heading, but is unable capture pitch and roll measurements. Northing and Easting position components are extracted in meters using the Universal-Transverse Mercator (UTM) python package\cite{pythonUTM}. The heading measurements are converted into unit quaternions for comparison in OpenVINS Evaluation toolkit.

OpenVINS and VINS-Fusion are selected for trajectory estimation of the collected data because they represent the dichotomy of strategies of VIO and are state-of-the-art implementations. OpenVINS uses a multi-state constraint sliding Kalman Filter (MSCKF) for estimation with Simulatenous Localization and Mapping (SLAM) feature tracking for loop-closure\cite{Geneva2020ICRA}. VINS-Fusion uses non-linear optimization based system for pose estimation with loop detection and pose graph optimization for global consistency\cite{qin2019general}. These algorithms have shown that they perform adequately across indoor environments\cite{Geneva2020ICRA}. The rt4 and rt5 configurations for the implementations of OpenVINS and VINS-Fusion are provided on the ROOAD website.

The OpenVINS Evaluation toolkit includes integration of the trajectory estimation performance metrics Absolute Trajectory Error (ATE) and Relative Pose Error (RPE)\cite{Geneva2020ICRA,Zhang18iros}.  The ATE simply measures the error between the given ground truth trajectory \(x_{k,i}\) and the estimated trajectory \(\hat{x}_{k,i}^+\) after the best alignment transformation has been performed for \(K\) pose measurements over \(N\) runs (exhibited in Fig.~\ref{fig:trajectoryerrortopdownupdown}). 

\[ e_{ATE}=\frac{1}{N}\sum_{i=1}^N\sqrt{\frac{1}{K}\sum_{k=1}^K\|x_{k,i}-\hat{x}_{k,i}^+\|_2^2} \]

ATE is a helpful metric when comparing the performance of one algorithm to another, but it is missing key insights of how the trajectory estimate is behaving for different measurement periods and traversal lengths. RPE measures how the estimation diverges from the groundtruth over trajectory segments with lengths \(D=[d_1,d_2,...,d_i]\). The relative pose error for \(D_i\) number of segments corresponding to length \(d_i\) is the average magnitude of the error between the groundtruth pose evolution \(\Tilde{x}_{k,i}\) and the estimated pose evolution \(\hat{\Tilde{x}}_{k,i}\), where the evolution measures the relative change in pose over length \(d_i\) for the \(i\)-th segment.
\[ \Tilde{x}_{k,i} = x_k - x_{k+d_i} \]

\[ e_{RPE,d_i}=\frac{1}{D_i}\sum_{k=1}^{D_i}\|\Tilde{x}_{k,i}-\hat{\Tilde{x}}_{k,i}\|_2\]

Using the ATE and RPE metrics, OpenVINS shows higher stability and consistency in our off-road scenario, but VINS-Fusion out-performs OpenVINS in certain segments. The Table~\ref{tbl:datasetrpe} shows the RPE performance of OpenVINS and VINS-Fusion side-by-side. VINS-Fusion has lower translational RPE in rt4 gravel and both up-down datasets and OpenVINS has lower orientation RPE in those datasets. OpenVINS out-performs VINS-Fusion on all other datasets. Particularly, the rim datasets demonstrate a challenging scenario for the algorithms, and VINS-Fusion demonstrates large estimation divergences. Over 24 meter segments, OpenVINS has minimum orientation and translation RPE of 2.7\textdegree~and 1.8 m, respectively, while VINS-Fusion achieves 3.1\textdegree~and 1.6 m, respectively. OpenVINS demonstrates 3.1-9.4x higher orientation and 9.1-37x higher position RPE and VINS-Fusion achieves (on non-diverging runs) 2-7.5x higher orientation and 11-30x higher position RPE than their corresponding segment length runs on the EuRoC dataset\cite{Geneva2020ICRA}. Fig.~\ref{fig:translationerror} exhibits the ATE for in the x, y, and z directions over the trajectories' length for OpenVINS and VINS-Fusion across the evening (rt4) datasets, and Fig.~\ref{fig:trajectoryerrortopdownupdown} shows the error between the aligned trajectories for the morning (rt5) datasets.

Many VIO research applications have turned to Kalibr as the de facto camera-IMU extrinsics estimator. To measure the efficacy of Kalibr camera-IMU extrinsic calibration, several calibration sequences are provided with measured camera linear and angular positions using the Edmund Optic's linear rail hatching and rotary stage orientations shown in Fig.~\ref{fig:camimuori}. The linear rail provides a Vernier scale with a resolution of 0.1 mm and the rotary stage provides a resolution of 1 degree. This position and orientation estimation of the camera-IMU extrinsics is performed using Kalibr's calibrate camera-IMU tool. Processing these sequences in Kalibr results in Table \ref{tbl:kalibr_pos} and \ref{tbl:kalibr_angle}. Kalibr shows strong repeatability and accuracy within \(\pm\)1\textdegree~for rotational measurements and \(\pm\)1 mm for the x and y measurements in the camera frame. The linear estimation of depth (z) axis of the camera is the least accurate with an average error of 47 mm and a measurement standard deviation of 10 mm. It is important to note that it is difficult to accurately hand measure the exact position of the aperture and height of the camera sensor with respect to the IMU's exact sensor location when these sensors are enclosed, which grants more confidence to Kalibr's results.
\section{Summary and Future Work}

In this paper, we present ROOAD, a collection of off-road monocular visual-inertial data sequences, to contribute to the research community. We evaluate two state-of-the-art VIO implementations, OpenVINS and VINS-Fusion, which report promising results in structured environments \cite{Geneva2020ICRA, qin2019general} but performed poorly in unstructured environment data sequences. OpenVINS exhibits consistent trajectory estimation performance across our data sequences when compared to VINS-Fusion which diverges heavily in our "rim" data sequences. VINS-Fusion does outperform OpenVINS in the positional estimation on the rt4 gravel and both of our up-down data sets, but OpenVINS out performs VINS-Fusion in all other scenarios. Both implementations demonstrate between 200-3000\% higher error on our off-road dataset when compared to the EuRoC dataset. OpenVINS demonstrates a maximum 40 m segment RPE of 9.4\textdegree~and 3.7 m while VINS-Fusion, when stable, achieves 9.5\textdegree~and 3.1 m.  This performance gap shows that there is not a direct transference of VIO capabilities from developed to off-road environments and that there is opportunity for further improvement. 

Additionally, we validate the efficacy of Kalibr camera-IMU extrinsics calibration tool by collecting multiple calibration datasets at different linear and angular camera-IMU offsets. Kalibr has exceeded our expectations with accuracy and repeatability within 1\textdegree~for our angular measurements and adequate estimation along its linear axes with the best of \(\pm\)1 mm on its x-axis and at worst \(\pm\)10 mm on z-axis in the camera frame. 

In the future, we consider extending our off-road data collection to include similarly PTP time-stamped stereo cameras and LIDAR(s). In order to capture more scenarios, we would like to attach this sensor suite to multiple vehicles for similar data collection runs with their different geometry, dynamics, and excitation profiles. We are also interested in the effects of varying exposure times in state estimation and allowing for automatic exposure control by the camera to adjust for the varied lighting conditions outdoors.




\bibliographystyle{IEEEtran}
\bibliography{IEEEabrv,references}

\end{document}